\documentclass[11pt]{article}
\usepackage[preprint]{acl}
\usepackage{times}
\usepackage{latexsym}
\usepackage[T1]{fontenc}
\usepackage[utf8]{inputenc}
\usepackage{microtype}
\usepackage{graphicx}
\newcommand{\subbreve}[1]{\ensuremath{\underset{\raisebox{0.35ex}{$\scriptscriptstyle\smile$}}{\text{#1}}}}
\usepackage{newunicodechar}
\newunicodechar{ḍ}{\d{d}}\newunicodechar{ḥ}{\d{h}}\newunicodechar{ḳ}{\d{k}}\newunicodechar{ṣ}{\d{s}}
\newunicodechar{ṭ}{\d{t}}\newunicodechar{ẓ}{\d{z}}\newunicodechar{ḫ}{\subbreve{h}}\newunicodechar{ẕ}{\b{z}}
\newunicodechar{Ṭ}{\d{T}}\newunicodechar{Ḥ}{\d{H}}\newunicodechar{Ḳ}{\d{K}}\newunicodechar{Ṣ}{\d{S}}\newunicodechar{Ẓ}{\d{Z}}
\newunicodechar{ʿ}{\textquoteleft}\newunicodechar{ʾ}{\textquoteright}
\newunicodechar{ˑ}{\textperiodcentered}\newunicodechar{‿}{\raisebox{-0.55ex}{$\scriptstyle\smile$}}
\newunicodechar{→}{$\rightarrow$}\newunicodechar{κ}{$\kappa$}\newunicodechar{×}{$\times$}
\usepackage{amsmath,amssymb}
\usepackage{booktabs,array}
\usepackage{tikz}
\usetikzlibrary{arrows.meta,positioning,calc,shapes.geometric}
\usepackage{xcolor}
\definecolor{cblue}{HTML}{2A78D6}\definecolor{corange}{HTML}{EB6834}\definecolor{caqua}{HTML}{1BAF7A}
\usepackage{enumitem}
\usepackage{url}
\setlist{nosep,leftmargin=1.4em}
\newcommand{\ot}[1]{\textit{#1}}

\newcommand{\slong}{\mbox{--}}\newcommand{\sshort}{\ensuremath{\scriptstyle\cup}}
\makeatletter\DeclareRobustCommand{\pat}[1]{\@tfor\pc:=#1\do{\if\pc-\slong\else\if\pc.\sshort\else\pc\fi\fi\hspace{0.12em}}}\makeatother
\usepackage{float}
\newcommand{\sym}[1]{\mathsf{#1}}
\newcommand{\Lw}{\sym{L}}\newcommand{\Sw}{\sym{S}}\newcommand{\Hw}{\sym{H}}\newcommand{\Xw}{\sym{X}}\newcommand{\Iw}{\sym{I}}\newcommand{\Kw}{\sym{K}}\newcommand{\Ow}{\sym{O}}
\newcommand{\BOS}{\textsc{bos}}\newcommand{\EOS}{\textsc{eos}}
\DeclareMathOperator*{\argmax}{arg\,max}
\makeatletter\newcommand{\num}[1]{\csname num@#1\endcsname}\makeatother
\expandafter\def\csname num@ABL_flat\endcsname{49.2}
\expandafter\def\csname num@ABL_lexicon\endcsname{40.0}
\expandafter\def\csname num@ABL_noem\endcsname{18.2}
\expandafter\def\csname num@ABL_nohmm\endcsname{44.2}
\expandafter\def\csname num@ABL_nosymbols\endcsname{42.4}
\expandafter\def\csname num@CLF_line\endcsname{92.0}
\expandafter\def\csname num@COSTOCR\endcsname{33}
\expandafter\def\csname num@COSTWEB\endcsname{18}
\expandafter\def\csname num@DET_ocr_prec\endcsname{66}
\expandafter\def\csname num@DET_ocr_rec\endcsname{54}
\expandafter\def\csname num@DET_web_prec\endcsname{51}
\expandafter\def\csname num@DET_web_rec\endcsname{42}
\expandafter\def\csname num@DMAX\endcsname{22.7}
\expandafter\def\csname num@DMIN\endcsname{10.8}
\expandafter\def\csname num@ERR_rest\endcsname{9.4}
\expandafter\def\csname num@ERR_spelling\endcsname{13.0}
\expandafter\def\csname num@ERR_total\endcsname{22.4}
\expandafter\def\csname num@FLASH_asked\endcsname{1,607}
\expandafter\def\csname num@FLASH_exact\endcsname{40.0}
\expandafter\def\csname num@FORMSLOSS\endcsname{6--14}
\expandafter\def\csname num@KKPLAIN_exact\endcsname{2.4}
\expandafter\def\csname num@KKPLAIN_poem\endcsname{91.0}
\expandafter\def\csname num@KKSCH_exact\endcsname{34.4}
\expandafter\def\csname num@KurtKararestorer_exact\endcsname{26.8}
\expandafter\def\csname num@KurtKararestorer_t1line\endcsname{84.1}
\expandafter\def\csname num@KurtKararestorer_t1poem\endcsname{98.9}
\expandafter\def\csname num@NGOLD\endcsname{500}
\expandafter\def\csname num@NOISEMETER\endcsname{91--95}
\expandafter\def\csname num@NTEST_ALL\endcsname{1,731}
\expandafter\def\csname num@NTEST_LINES\endcsname{15,228}
\expandafter\def\csname num@NTEST_OUT\endcsname{124}
\expandafter\def\csname num@NTEST_POEMS\endcsname{1,607}
\expandafter\def\csname num@OPUS_exact\endcsname{63.6}
\expandafter\def\csname num@OURS_exact\endcsname{77.6}
\expandafter\def\csname num@OURS_t1line\endcsname{98.2}
\expandafter\def\csname num@RES_right\endcsname{93.5}
\expandafter\def\csname num@RES_unseen\endcsname{59}
\expandafter\def\csname num@RES_words\endcsname{113,195}
\expandafter\def\csname num@ROB_ocr_KK_exact\endcsname{7.8}
\expandafter\def\csname num@ROB_web_KK_exact\endcsname{18.6}
\expandafter\def\csname num@SCH_HMM\endcsname{90.6}
\expandafter\def\csname num@SIG_opus_diff\endcsname{+14.0}
\expandafter\def\csname num@SIG_opus_hi\endcsname{+18.6}
\expandafter\def\csname num@SIG_opus_lo\endcsname{+9.4}
\expandafter\def\csname num@SIG_opus_ptxt\endcsname{p<0.001}
\expandafter\def\csname num@TAG_exact\endcsname{55.0}
\expandafter\def\csname num@EM_K_long_ar\endcsname{.002}
\expandafter\def\csname num@EM_K_long_tr\endcsname{$3{\cdot}10^{-4}$}
\expandafter\def\csname num@EM_imale_ar\endcsname{13}
\expandafter\def\csname num@EM_imale_tr\endcsname{28}
\expandafter\def\csname num@EM_zihaf_ar\endcsname{1.1}
\expandafter\def\csname num@EM_zihaf_tr\endcsname{0.04}

\expandafter\def\csname num@N_METERS\endcsname{14}
\expandafter\def\csname num@N_TEMPLATES\endcsname{17}
\expandafter\def\csname num@N_TRAIN_INVENTORY\endcsname{1,511}

\expandafter\def\csname num@ROBX_ocr_factor_damage\endcsname{0.91}
\expandafter\def\csname num@ROBX_ocr_factor_forms\endcsname{1.15}
\expandafter\def\csname num@ROBX_ocr_sources\endcsname{6}
\expandafter\def\csname num@ROBX_ocr_survey_misras\endcsname{13,532}
\expandafter\def\csname num@ROBX_ocr_survey_words\endcsname{90,855}
\expandafter\def\csname num@ROBX_ocr_wordedits\endcsname{1946}
\expandafter\def\csname num@ROBX_web_beyit_left_out\endcsname{10.5}
\expandafter\def\csname num@ROBX_web_factor_damage\endcsname{1.25}
\expandafter\def\csname num@ROBX_web_factor_forms\endcsname{2.83}
\expandafter\def\csname num@ROBX_web_intensity_max\endcsname{1.73}
\expandafter\def\csname num@ROBX_web_intensity_min\endcsname{0.33}
\expandafter\def\csname num@ROBX_web_sources\endcsname{33}
\expandafter\def\csname num@ROBX_web_survey_misras\endcsname{512}
\expandafter\def\csname num@ROBX_web_survey_words\endcsname{3,579}
\expandafter\def\csname num@ROBX_web_wordedits\endcsname{126}

\makeatletter\newcommand{\err}[1]{\csname err@#1\endcsname}\makeatother
\expandafter\def\csname err@ours_spelling\endcsname{17}
\expandafter\def\csname err@ours_vowellength\endcsname{63}
\expandafter\def\csname err@ours_medd\endcsname{2}
\expandafter\def\csname err@ours_linking\endcsname{30}
\expandafter\def\csname err@ours_total\endcsname{112}

\title{A Lattice-Based HMM Approach to Aruz Scansion of Ottoman Poetry}

\author{Mehmet Emre Tiryaki \\ Johns Hopkins University \\ \texttt{mtiryak1@jhu.edu}}

\begin{document}
\maketitle

\begin{abstract}
Ottoman Divan poetry is composed in \emph{aruz}, a quantitative meter based on the weights of syllables. The meter is sensitive to specific characters recorded in the original writing in the Arabic script. This work is concerned with automatic meter finding and scansion alignment for Ottoman Divan poetry, especially on modern Turkish spelling without the special diacritics that scholarly transliterations of Ottoman preserve. We define two tasks on this input, first identifying the meter used in the poem, and second aligning the meter’s weights with the syllabification of the poem, identifying the flaws and freedoms the poet has taken along the way. We build a corpus of 52,000 lines split by poet, with scansion annotations for \num{NGOLD} test lines, and a robustness benchmark generated at the deviation rates measured on real web and OCR copies. Our model is a hidden Markov model over the positions of each meter, decoded over a lattice of every candidate path, and trained by EM from the printed meter, without scansion labels. On a held-out set it scans \num{OURS_exact}\% of the test lines fully correctly, against \num{KurtKararestorer_exact}\% for the available rule-based scanner, \num{TAG_exact}\% for a neural tagger trained with the same supervision and \num{OPUS_exact}\% for Claude Opus 5.5 few-shot. The code is open-sourced and the tool is released on the web.

\end{abstract}

\section{Introduction}\label{sec:intro}

Scansion, the task of assigning a rhythmic structure to a written line of
poetry, is a sequence labeling problem that NLP has approached with rules,
HMMs, CRFs, neural taggers and LLMs
\citep{agirrezabal2016machine,agirrezabal2017comparison,ruizfabo2026scansion}. One type of verse
that has received relatively little attention in this context is Ottoman Divan poetry, the
classical Turkish poetry of the educated elite, written from the
late thirteenth century to the second half of the nineteenth century. Divan poetry is composed in a
quantitative meter called \emph{aruz}, taken over from
Arabic and Persian poetry \citep{cetin1991aruz,ipekten2010aruz}. In aruz every
syllable is either open or closed, closed if its vowel is long or it ends
in a consonant, open otherwise. Each meter fixes a sequence of short and
long positions, and a line fills it with open and closed syllables. In this paper, we address
the scansion of Divan poetry: identifying a poem's meter and aligning each
syllable of each line to its position in that meter.

An automated tool that can robustly determine a Divan poem's meter and align
each syllable has many uses for Turkish literature researchers and
poetry readers. The meter directly influences prosody, since a reader who is not aware of the correct meter cannot recite the line correctly. Finding it by hand is slow and
error-prone even for trained readers, who can disagree on the same
poem~\citep{kurt2012arud}, and it is a standing difficulty for students of
Turkish literature. Editors depend on the scansion, since a line that will not scan is the sign of a
misreading or a misprint~\citep{koksal2009metin}. And more than a thousand divan collections remain in
Turkish libraries, many of them unstudied~\citep{kurt2012arud}. Tracing the meters, flaws
and choices of poets across eras would benefit greatly from an automated
scanner.

The problem is not trivial, for two main reasons. The first is the script. The
poems were written in the Arabic alphabet, which records the long vowels and
consonants the meter counts. Since the alphabet reform of 1928~\citep{lewis1999turkish},
Turkish has been written in Latin letters, and the original texts, in a script few now
read and hard to OCR, remain largely
without machine-readable text~\citep{kirmizialtin2020automated}. Scholarly
editions mark modern Latin with diacritics to signify differences in the original writing, but most readers use this
poetry in plain modern Turkish spelling, in school, on websites or
typed from a manuscript, without special diacritics
(\S\ref{sec:aruz}). A scanner must therefore work from modern spelling, where
the weights have to be restored, and should not depend on the scholarly marks
even when they are present, since editions do not always supply them and can be incorrect or inconsistent. Here is
a couplet by the fifteenth-century poet Mihrî Hatun in the Arabic script in
which it was written, in the scholarly transcription of its edition, and in
the modern spelling%
\footnote{\citet{arslan2018mihri}, poem 193, p.~152, couplet 7; both lines are in the annotated test set
(\S\ref{sec:data}).}
\begin{center}
{\small\setlength{\tabcolsep}{4pt}%
\begin{tabular}{@{}l@{\hspace{6pt}}l@{}}
original & \raisebox{-1.6em}{\includegraphics[height=2.4em]{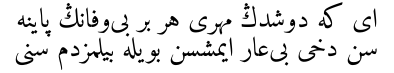}}\\[1.4em]
scholarly & \ot{Ey ki düşdüñ Mihrī her bir bī-vefānuñ pāyına}\\
          & \ot{Sen daḫi bī-ʿār imişsin böyle bilmezdüm seni}\\[3pt]
modern    & \ot{Ey ki düşdün Mihri her bir bi-vefanun payına}\\
          & \ot{Sen dahi bi-ar imişsin böyle bilmezdüm seni}\\
\end{tabular}}\\[3pt]
{\small (``You who fell, Mihrî, at the feet of every ingrate / so you too are
shameless. I did not know you so.'')}
\end{center}

Take \ot{bi-vefanun payı}, `at the feet of every ingrate', from the first
line. The meter requires its six syllables to be long, short, long, long, long,
short:
\begin{center}
\small
\begin{tabular}{@{}l*{6}{c}@{}}
meter   & \pat{-} & \pat{.} & \pat{-} & \pat{-} & \pat{-} & \pat{.}\\
scholarly & \ot{bī} & \ot{ve} & \ot{fā} & \ot{nuñ} & \ot{pā} & \ot{yı}\\
modern  & \ot{bi} & \ot{ve} & \ot{fa} & \ot{nun} & \ot{pa} & \ot{yı}\\
\end{tabular}
\end{center}
\ot{bī}, \ot{fā} and \ot{pā} are closed because their vowels are long, and the
scholarly edition marks the long vowel. The modern spelling does not, so \ot{bi},
\ot{fa} and \ot{pa} look exactly like the open \ot{ve} and \ot{yı}.

The second reason is that, even when the original letters are known, a line does not always
match its meter syllable for syllable. Aruz tolerates a set of departures from
the expected weights, called flaws
(\emph{kusur})~\citep{duzenli2018aruz}. For example an open syllable may fill a long
position, a closed one a short position, and a word may or may not be linked
to the next (\S\ref{sec:aruz}). In the couplet above, the open second syllable
of \ot{dahi} fills a long position and is read with a long vowel. These are not marked in the text, and which flaws a line takes is
determined by the meter of the poem, not by the line itself, so a single line
admits many readings.

These decisions cannot be made syllable by syllable. Whether a syllable takes
a flaw depends on the syllables before and after it, and on the meter, which
is not given. Since a single line often fits more than one meter, the meter,
and with it the reading of every line, can only be settled for the poem as a
whole.

Computational work on aruz has concentrated on Arabic
(\S\ref{sec:related}). Rule-based systems~\citep{abuata2018rule} and recurrent
networks~\citep{yousef2019learning,alshaibani2020meter} identify the meter of
a verse and they work on Arabic script. For Ottoman poetry the only system we know
of is that of \citet{kurt2012arud}, which is a deterministic rule-based algorithm running on scholarly edition strings. It is bound by the scholarly
transliteration. It reads the syllable weights directly off the edition's
letters, so it cannot be used on the modern spelling, and
even on scholarly text a long-vowel mark the edition left out
is recognized as a flaw instead of being corrected. This over-dependence on the scholarly edition input makes the system prone to mistakes. Large language models, used zero- or few-shot, are the obvious
alternative. As we show in \S\ref{sec:results}, the strongest models we tried,
prompted with the rules, the meters and worked examples, read at best
\num{OPUS_exact}\% of held-out lines correctly.

We first define two tasks on a poem given in plain modern Turkish spelling
with no meter marked, a form to which a scholarly transcription reduces by
removing its diacritics, identifying the poem's meter, and scanning each of
its lines with every syllable's position in the meter and any flaw it takes.
For these tasks we build a corpus of 52,000 lines from thirteen critical
editions, split by poet, and scansion annotations for \num{NGOLD} test lines
(\S\ref{sec:tasks}, \S\ref{sec:data}).

Our model follows from the difficulties above. A trained spelling model proposes
scholarly forms for each word, and every reading of the line these forms allow
is laid out in a lattice. A hidden Markov model over the positions of the meters
scores the paths through that lattice. The only supervision is
the printed meter of each training poem, from which the model is trained by EM
(\S\ref{sec:model}). We compare it with the scanner of \citet{kurt2012arud} run on
the same spelling model, a neural tagger trained with the same
supervision, and with two large language models, and we examine which parts of the model
account for the difference (\S\ref{sec:baselines}, \S\ref{sec:results}).
Since the text people actually scan is rarely clean edition text, we also
build a perturbation benchmark at the deviation rates measured on web and OCR
copies and test every system but the language models on it (\S\ref{sec:data}, \S\ref{sec:results}). The code and data pipeline are released at
\url{https://github.com/MehmetETiryaki/aruz-paper}, and the model is freely available as a web tool at
\url{https://mehmetemretiryaki.com/aruz}.

\section{Aruz}\label{sec:aruz}

\paragraph{Syllable weight.}
Aruz counts syllables by weight, not by stress. A syllable is \emph{open} if
it ends in a short vowel, as \ot{ka}, and \emph{closed} if it ends in a long
vowel or a consonant, as \ot{kā} or \ot{kal}. Some syllables, those with a
long vowel followed by a consonant, as \ot{kār}, or a short vowel followed by
two consonants, as \ot{kalb}, hold the place of a closed syllable followed by
an open one; these are
\emph{overlong} syllables (\emph{medli}), and the reading is called \emph{medd}. The last syllable of a line counts as closed regardless
of its letters~\citep{ipekten2010aruz}.

The Arabic alphabet writes the long vowels and the consonants, so the weight
of a syllable can be read from the original once the word is known~\citep{bugday2009routledge}. Scholarly editions keep this
in their transcription, with a macron on long vowels (\ot{ā}, \ot{ī},
\ot{ū}) and the signs \ot{ʿ} and \ot{ʾ} for \emph{ayn} and \emph{hemze}, two
consonants Turkish does not pronounce but the meter counts. The modern
spelling has neither, and marks a long vowel with a circumflex only where the
word would otherwise be confused with another~\citep{lewis1999turkish,yesiloz2021uzun}.

Syllabification runs across word boundaries as in ordinary
Turkish speech. A word-final consonant before a vowel-initial word is
resyllabified as the onset of the next syllable (liaison, Turkish
\emph{ulama})~\citep{goksel2005turkish}. For example \ot{bir elma} `an apple' is syllabified \ot{bi-rel-ma}.

\paragraph{Feet and meters.}
A meter is a fixed sequence of feet (\emph{tefʿile}), each named by a
mnemonic word whose own syllables have the foot's
weights~\citep{cetin1991aruz}. A poem is written in a single meter, and every
line of it must fill the same sequence. We write a long position \slong{} and
a short one \sshort{}; a closed syllable fills a long position and an open one
a short position. The first line of the couplet fills its meter, \emph{remel},
as follows:
\begin{center}
\small
\begin{tabular}{@{}llll@{}}
fâʿilâtün & fâʿilâtün & fâʿilâtün & fâʿilün \\
\pat{-.--} & \pat{-.--} & \pat{-.--} & \pat{-.-} \\
\ot{Ey ki düş düñ} & \ot{Mih rī her bir} & \ot{bī ve fā nuñ} & \ot{pā yı na} \\
\end{tabular}
\end{center}
Scanning a line means producing this alignment, each syllable assigned to one
position of the meter. Foot boundaries need not fall at word boundaries, as
inside \ot{düşdüñ} and \ot{bī-vefānuñ} above. Ottoman poetry uses a few dozen
meters, of which a handful account for most poems~\citep{kurt2012arud}.

\paragraph{Flaws.}
Poets often depart from the rules above to fit the words and meaning they
want into the meter. A flaw (\emph{kusur}) is a reading in which a syllable
does not fill the slot that proper application of the rules requires. Four main flaws are discussed. In
\emph{imâle}, the most common flaw, an open syllable fills a long position and
is read with a long vowel, as the Turkish \ot{daḫi} is read \ot{daḫī} in the
second line of the couplet. In \emph{zihaf}, a closed syllable fills a short
position and its long vowel is read short, as the pen-name \ot{Mihrī} is in the first. A line that reads an overlong syllable in one position (\emph{no
medd}), or keeps a final consonant before a following vowel (\emph{no
linking}), also counts as a flaw. Which flaws are
tolerated and which are taken as the mark of an unskilled poet depends on the
origin of the word, the period of the poem and the standing of the
poet~\citep{duzenli2018aruz,sarac2022divan}. In principle, however, a poet may take any of them anywhere.
None of them is usually marked in any spelling~\citep{duzenli2018aruz}.

\section{Tasks and Evaluation}\label{sec:tasks}

As discussed in \S\ref{sec:intro}, most poems are available in
modern Turkish spelling. A scholarly transcription, where
there is one, can be normalized to the same spelling. Taking its marks as given
would carry the marks the edition left out or got wrong into the scansion as flaws. Both tasks therefore take a poem in plain modern
spelling with no meter given.

\textbf{Task 1}, meter identification, is to name the meter of a group of
lines from an inventory of fourteen (Appendix~\ref{app:meters}). We report
the two cases that matter in practice, a whole poem and a single line; the
second is harder, since a single line often fits more than one meter.

\textbf{Task 2}, scansion, is to give for every line each syllable as it is read, the position or positions it
fills (\slong{} long, \sshort{} short, \slong\sshort{} for medd), and its flaw: imâle, zihaf, no medd, no linking,
or none. Figure~\ref{fig:scansion} shows the reference scansion
of the couplet of \S\ref{sec:intro}.
\begin{figure}[t]\centering
\setlength{\tabcolsep}{1.5pt}\small
\resizebox{\columnwidth}{!}{%
\begin{tabular}{@{}l*{15}{c}@{}}\toprule
input & \ot{Ey} & \ot{ki} & \ot{düş} & \ot{dün} & \ot{Mih} & \ot{ri} & \ot{her} & \ot{bir} & \ot{bi} & \ot{ve} & \ot{fa} & \ot{nun} & \ot{pa} & \ot{yı} & \ot{na}\\
read & \ot{Ey} & \ot{ki} & \ot{düş} & \ot{düñ} & \ot{Mih} & \ot{ri} & \ot{her} & \ot{bir} & \ot{bī} & \ot{ve} & \ot{fā} & \ot{nuñ} & \ot{pā} & \ot{yı} & \ot{na}\\
pos. & \slong{} & \sshort{} & \slong{} & \slong{} & \slong{} & \sshort{} & \slong{} & \slong{} & \slong{} & \sshort{} & \slong{} & \slong{} & \slong{} & \sshort{} & \slong{}\\
flaw &  &  &  &  &  & zihaf &  &  &  &  &  &  &  &  & \\
\midrule
input & \ot{Sen} & \ot{da} & \ot{hi} & \ot{bi} & \ot{a} & \ot{ri} & \ot{miş} & \ot{sin} & \ot{böy} & \ot{le} & \ot{bil} & \ot{mez} & \ot{düm} & \ot{se} & \ot{ni}\\
read & \ot{Sen} & \ot{da} & \ot{ḫī} & \ot{bī} & \ot{ʿā‿} & \ot{ri} & \ot{miş} & \ot{sin} & \ot{böy} & \ot{le} & \ot{bil} & \ot{mez} & \ot{düm} & \ot{se} & \ot{ni}\\
pos. & \slong{} & \sshort{} & \slong{} & \slong{} & \slong{} & \sshort{} & \slong{} & \slong{} & \slong{} & \sshort{} & \slong{} & \slong{} & \slong{} & \sshort{} & \slong{}\\
flaw &  &  & imâle &  &  &  &  &  &  &  &  &  &  &  & \\
\bottomrule
\end{tabular}}

\caption{Task 2 output for the couplet of \S\ref{sec:intro}, the reference scansion: the input in modern letters, each
syllable as it is read, the position it fills (\slong{} long, \sshort{} short, \slong\sshort{} medd) and its flaw,
one of imâle, zihaf, no medd, no linking or none. \ot{‿} after a syllable marks a linking, its consonant read
with the next word.}
\label{fig:scansion}
\end{figure}

\paragraph{Metrics.}
Task 1 is scored by accuracy over the fourteen meters, per poem and per line. For Task 2, we use several metrics. The main measure is
\emph{full-line accuracy}, the share of lines whose output equals the reference in every syllable, position, flaw and
word-boundary decision. \emph{Syllable F1} is F1 over syllable spans, and \emph{flaw F1} is the pooled F1 score of
each flaw given a syllable.

\section{Data}\label{sec:data}

\paragraph{Corpus.}
The corpus is thirteen critical editions of Divan poets of the fifteenth to the eighteenth century, taken from
LATOC~\citep{yilandiloglu2026latoc}, a corpus of 143 expert-transliterated Ottoman works (13 million words) that
comes with the tools to extract and structure the text from the publicly available PDFs, and hand-corrected where
the extraction had failed. The split is by poet, so that no poet's vocabulary or metrical habits are shared between training and evaluation. The input is the editions' transcription with its marks removed. This
gives a training set of six poets, 1,644 poems and 18,282 lines; a development set of four poets, 1,133 poems and
18,090 lines; and a test set of three poets, 1,731 poems and 15,992 lines (Table~\ref{tab:data}). Evaluation uses
the \num{NTEST_POEMS} test poems (\num{NTEST_LINES} lines) whose printed meter is in the inventory.
\begin{table}[t]\centering\footnotesize\setlength{\tabcolsep}{3pt}
\begin{tabular}{lp{3.6cm}rr}\toprule
split & poets (century) & poems & lines\\\midrule
train & Bahtî (17), Sürûrî (16), Dürrî (18), Esrâr (18), Nâfiz (18), Hâmî (18) & 1,644 & 18,282\\
dev & Tâcîzâde (15), Yahyâ (16), Şehrî (17), Vehbî (18) & 1,133 & 18,090\\
test & Fuzûlî (16), Mihrî Hatun (15), Murâdî (16) & 1,731 & 15,992\\\bottomrule
\end{tabular}
\caption{The corpus, split by poet.}\label{tab:data}
\end{table}

\paragraph{Scansion annotation.}
To our knowledge, no scansion annotation exists publicly for these poems. So the annotation was created for this paper using an ensemble of three language models, Claude Opus 5.5, ChatGPT 6 and Gemini 3.1 Pro, all with medium thinking. The annotators were given
the poems' scholarly transcription, the gold meter, a guide on how to label, and few-shot examples. Note that this is not the task defined above.
With the scholarly transliteration and the gold meter the scansion is largely determined, and what remains to decide is the linking, the
medd and the flaw labels. No system in Table~\ref{tab:main-test} sees either of these, so a language model evaluated as a system in
\S\ref{sec:results} faces a harder task than it did as annotator.

We annotated \num{NGOLD} lines of the test set, whole couplets with at most one per poem, and 100 lines of the
development set. Per-syllable agreement is high (Cohen's κ~\citep{cohen1960coefficient} 0.959--0.978 pairwise, Fleiss'
κ~\citep{fleiss1971measuring} 0.969 over the 450
of the 500 test lines all three answered validly); whole-line agreement is 66--79\% pairwise. If two or more
valid answers agree (478 lines) the annotation is accepted as gold. The author, who reads aruz, decided the remaining 22 lines by hand and audited 200 of the lines the annotators had settled,
stratified by agreement and by the rare flaws, accepting 95.5\% of them (95\% CI 93.0--98.0).

\paragraph{Robustness benchmark.}
The text a user submits is rarely a scholarly transliteration with its marks removed. To measure how real copies
deviate from the editions, we collected \num{ROBX_web_sources} web pages and
\num{ROBX_ocr_sources} OCR'd books that copy corpus poems (\num{ROBX_web_survey_misras} and
\num{ROBX_ocr_survey_misras} lines), aligned them to the editions, and counted every kind of deviation
and its rate in the standard Turkish alphabet (Table~\ref{tab:rob-catalogue}, Appendix~\ref{app:robustness}). We then produce two noisy versions of the test set, a web copy and an OCR'd
book, by applying the observed edits to the clean lines at their measured
rates~\citep{belinkov2018synthetic,naplava2021understanding}, scaled so that each version reaches the measured
severity (Table~\ref{tab:rob-severity}), and two more at twice every rate. The reference scansion is unchanged.

\section{Model}\label{sec:model}

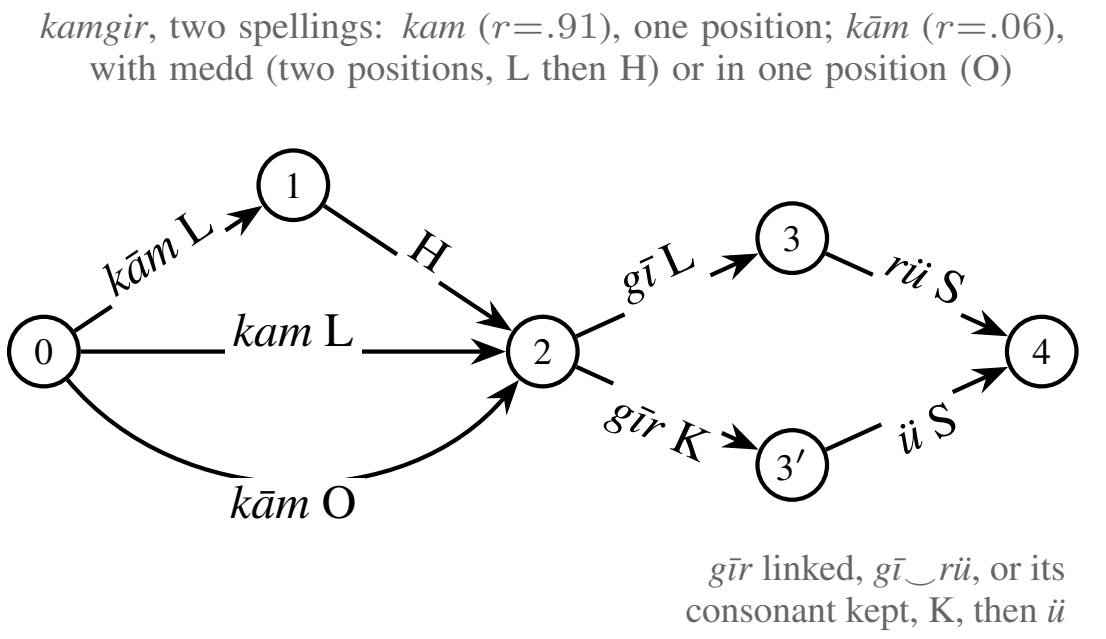
\begin{figure}[t]\centering
\begin{tikzpicture}[>=Stealth, font=\footnotesize,
  nd/.style={circle, draw, thick, fill=white, minimum size=4.6mm, inner sep=0pt, font=\scriptsize},
  syl/.style={font=\footnotesize, inner sep=1.5pt, fill=white},
  fork/.style={font=\scriptsize, color=black!60, inner sep=1pt, align=center}]
\node[nd] (n0) at (0,0) {0};
\node[nd] (n1) at (1.65,1.1) {1};
\node[nd] (n2) at (3.3,0) {2};
\node[nd] (n3a) at (4.95,0.75) {3};
\node[nd] (n3b) at (4.95,-0.75) {3$'$};
\node[nd] (n4) at (6.6,0) {4};
\draw[->, thick] (n0) -- (n1) node[syl, midway, sloped, above=-1pt] {\ot{kām}\ L};
\draw[->, thick] (n1) -- (n2) node[syl, midway, sloped, above=-1pt] {H};
\draw[->, thick] (n0) -- (n2) node[syl, midway, above=-1pt] {\ot{kam}\ L};
\draw[->, thick] (n0) to[out=-50, in=-130] node[syl, midway, below=-1pt] {\ot{kām}\ O} (n2);
\draw[->, thick] (n2) -- (n3a) node[syl, midway, sloped, above=-1pt] {\ot{gī}\ L};
\draw[->, thick] (n3a) -- (n4) node[syl, midway, sloped, above=-1pt] {\ot{rü}\ S};
\draw[->, thick] (n2) -- (n3b) node[syl, midway, sloped, below=-1pt] {\ot{gīr}\ K};
\draw[->, thick] (n3b) -- (n4) node[syl, midway, sloped, below=-1pt] {\ot{ü}\ S};
\node[fork, anchor=south, text width=6.9cm] at (3.35,1.7) {\ot{kamgir}, two spellings: \ot{kam} ($r{=}.91$), one position; \ot{kām} ($r{=}.06$), with medd (two positions, L then H) or in one position (O)};
\node[fork, anchor=north east, align=right] at (6.8,-1.3) {\ot{gīr} linked, \ot{gī‿rü}, or its\\ consonant kept, K, then \ot{ü}};
\end{tikzpicture}
\caption{The lattice of the foot \ot{Kamgir ü}. Nodes are the points between
syllables and each arc is one syllable with its label; a path from node 0 to
node 4 is one reading of the foot. \ot{kamgir} has two candidate spellings,
with the restorer's probability $r$ of each (its remaining candidates are not
drawn), and \ot{kām} and \ot{gīr} each give two alternatives, as described in
the text.}
\label{fig:lattice}
\end{figure}

\begin{figure}[t]\centering
\begin{tikzpicture}[>=Stealth, font=\footnotesize,
  sq/.style={draw, thick, minimum size=8mm, inner sep=0pt, font=\normalsize},
  ins/.style={diamond, draw=corange, thick, fill=corange!15, minimum size=6mm, inner sep=0pt, font=\scriptsize},
  em/.style={font=\scriptsize, anchor=north, align=left, inner sep=1pt}]
\node[sq] (q0) at (0,0) {x};
\node[sq] (q1) at (1.65,0) {$\cup$};
\node[sq] (q2) at (3.3,0) {--};
\node[sq] (q3) at (4.95,0) {--};
\node[font=\scriptsize, anchor=east] (b) at (-0.65,0) {\textsc{bos}};
\node[font=\scriptsize, anchor=west, align=left] (e) at (5.6,0) {next\\ foot};
\draw[->, thick] (b) -- (q0); \draw[->, thick] (q3) -- (e);
\foreach \a [evaluate=\a as \c using int(\a+1)] in {0,1,2} {
  \draw[->, thick] (q\a) -- (q\c);
  \node[ins] (i\a) at ($(q\a)!0.5!(q\c)+(0,1.0)$) {I};
  \draw[->, corange] (q\a.north) to[out=75, in=180] (i\a);
  \draw[->, corange] (i\a) to[out=0, in=105] (q\c.north);
}
\draw[->, dashed, corange, thick] (q0.south) to[out=-35, in=-145] (q2.south);
\draw[->, dashed, corange, thick] (q1.south) to[out=-35, in=-145] (q3.south);
\node[font=\scriptsize, anchor=south, color=black!60, align=center] at (2.45,1.45) {on to the next position $A=.997$; insert (I): a syllable the template\\ has no place for, $.001$; skip (dashed): a syllable missing, $.002$};
\foreach \a/\s in {0/{\ot{kām}\ L}, 1/{H}, 2/{\ot{gī}\ L}, 3/{\ot{rü}\ S}} {
  \draw[->, cblue, thick] (q\a.south) -- ++(0,-0.95);
  \node[font=\footnotesize, anchor=north, cblue] at ($(q\a.south)+(0,-1.0)$) {\s};
}
\node[font=\scriptsize, anchor=east, color=black!60, align=right] at (-0.65,-1.2) {emits};
\node[em] at ($(q1.south)+(0,-1.75)$) {a short position ($\cup$) emits\\ S\quad .95\\ H\quad .04\quad medd\\ L\quad .005\quad zihaf};
\node[em] at ($(q3.south)+(0,-1.75)$) {a long position (--) emits\\ L\quad .79\\ S\quad .21\quad imâle\\ K\quad .002\quad no linking\\ O\quad $6{\cdot}10^{-6}$\quad no medd};
\end{tikzpicture}
\caption{The meter HMM for the first foot of the poem's meter, whose first
position may be long or short (x). There is a match state per position, an
insert state after each, and a skip over each. Below the states, the reading
of Figure~\ref{fig:lattice} as the labels each position emits. The rates are learned by EM from the printed meters alone.}
\label{fig:hmm}
\end{figure}
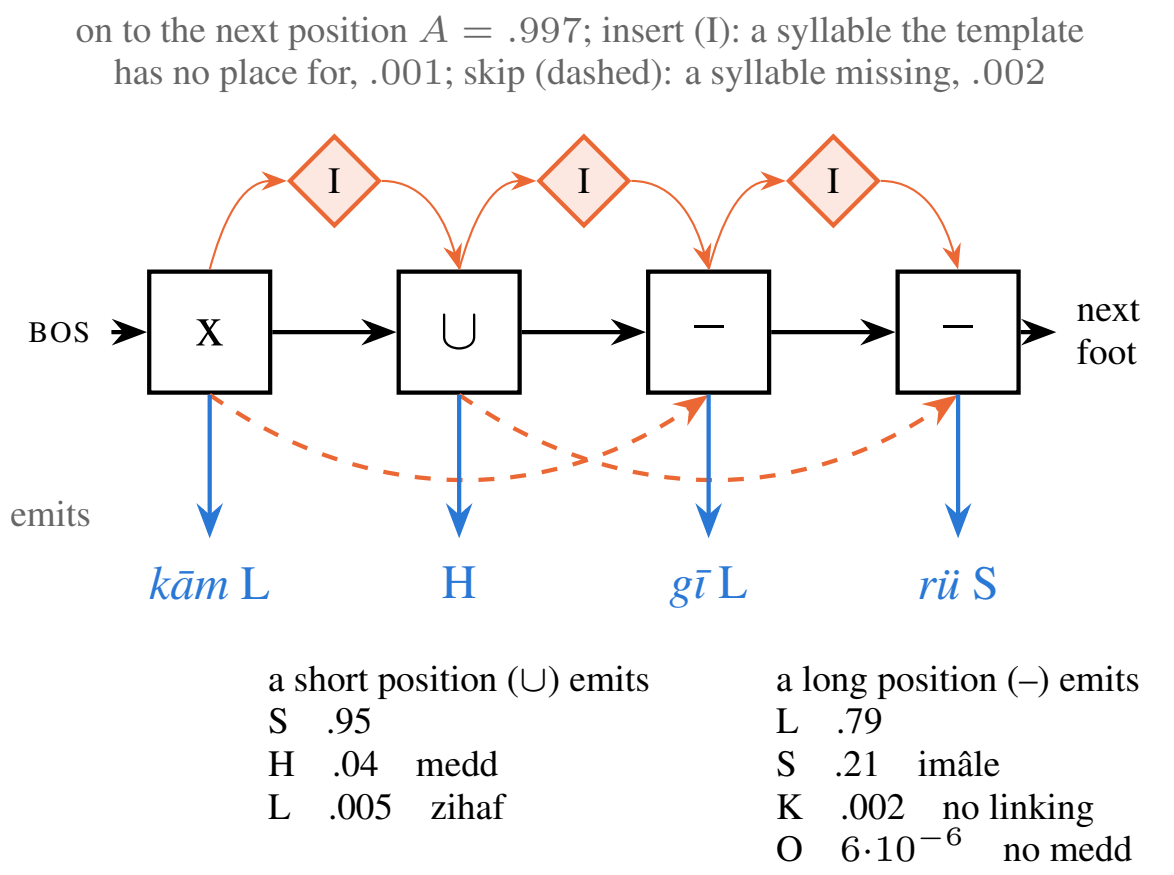

\begin{figure}[t]\centering
\begin{tikzpicture}[>=Stealth, font=\footnotesize,
  dot/.style={circle, fill=black!25, inner sep=0pt, minimum size=3pt},
  st/.style={circle, draw=black, thick, fill=white, minimum size=4.8mm, inner sep=0pt, font=\scriptsize},
  best/.style={cblue, line width=1.3pt},
  alt/.style={corange, line width=1.3pt, dashed},
  lab/.style={font=\scriptsize, inner sep=1pt, fill=white}]
\def\dx{1.15}\def\dy{0.82}
\foreach \c in {1,...,4} { \foreach \k in {1,...,4} { \node[dot] at (\c*\dx,-\k*\dy) {}; } }
\node[st] (s0) at (0,0) {\textsc{b}};
\node[font=\scriptsize, anchor=east, color=black!60] at (-0.45,0) {start};
\foreach \k/\t in {1/{1 (x)},2/{2 ($\cup$)},3/{3 (--)},4/{4 (--)}} { \node[font=\scriptsize, anchor=east, color=black!60] at (-0.45,-\k*\dy) {\t}; }
\node[font=\scriptsize, anchor=south east, color=black!60, align=right] at (-0.45,0.35) {position\\[-1pt] of the foot};
\foreach \c in {0,...,4} { \node[font=\scriptsize, anchor=north, color=black!60] at (\c*\dx,-4*\dy-0.3) {\c}; }
\node[font=\scriptsize, anchor=north, color=black!60] at (2*\dx,-4*\dy-0.62) {node of the lattice (text read so far)};
\draw[alt] (s0) -- (2*\dx,-1*\dy);
\draw[alt] (2*\dx,-1*\dy) -- (3*\dx,-3*\dy);
\draw[alt, transform canvas={yshift=-2.2pt}] (3*\dx,-3*\dy) -- (4*\dx,-4*\dy);
\node[lab, text=corange, anchor=south] at (1.55*\dx,-0.4*\dy) {\ot{kam}\,L};
\node[lab, text=corange, anchor=west] at (2.56*\dx,-2.0*\dy) {\ot{gī}\,L, a skip};
\draw[best] (s0) -- (1*\dx,-1*\dy);
\draw[best] (1*\dx,-1*\dy) -- (2*\dx,-2*\dy);
\draw[best] (2*\dx,-2*\dy) -- (3*\dx,-3*\dy);
\draw[best] (3*\dx,-3*\dy) -- (4*\dx,-4*\dy);
\node[lab, text=cblue, anchor=north east] at (0.55*\dx,-0.55*\dy) {\ot{kām}\,L};
\node[lab, text=cblue, anchor=north east] at (1.5*\dx,-1.6*\dy) {H};
\node[lab, text=cblue, anchor=north east] at (2.52*\dx,-2.58*\dy) {\ot{gī}\,L};
\node[lab, text=cblue, anchor=north east] at (3.52*\dx,-3.58*\dy) {\ot{rü}\,S};
\node[st] at (1*\dx,-1*\dy) {x}; \node[st] at (2*\dx,-1*\dy) {x}; \node[st] at (2*\dx,-2*\dy) {$\cup$};
\node[st] at (3*\dx,-3*\dy) {--}; \node[st] (e) at (4*\dx,-4*\dy) {--};
\node[font=\scriptsize, anchor=west, align=left] at ($(e.east)+(0.12,0)$) {$\log p$\\[-1pt] \textcolor{cblue}{$-31.5$}\\[-1pt] \textcolor{corange}{$-31.7$}};
\end{tikzpicture}
\caption{The HMM of Figure~\ref{fig:hmm} run over the lattice of
Figure~\ref{fig:lattice}. Each circle is a position of the foot (row) paired
with a node of the lattice (column); the arc read at each step is written on
it with its label. The blue path reads \ot{kām} with medd and \ot{gī‿rü}, one
position per step. The orange path reads the restorer's \ot{kam}, which
fills one position only, so \ot{gī} has to move two rows down and the second
position is skipped. The scores at the end node are log-probabilities; the
blue reading is chosen with a higher likelihood although the restorer preferred \ot{kam}, the orange reading. The two readings
that keep the consonant of \ot{gīr} score $-37.0$ and $-37.2$ and are not
drawn.}
\label{fig:trellis}
\end{figure}
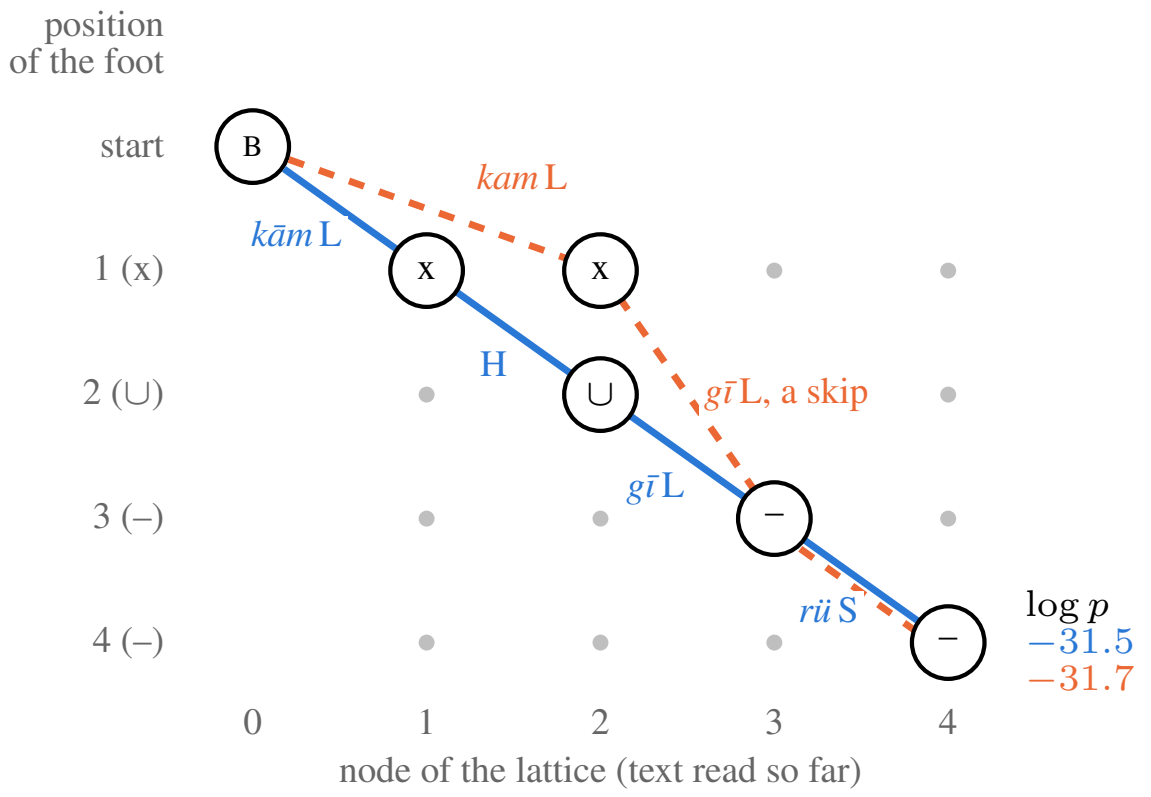

The model has three parts: a spelling model that proposes scholarly forms for each plain word, a lattice that holds
every reading of the line those forms allow, and a hidden Markov model over the positions of the meters that scores
the paths of the lattice. The meter of a poem is the one with the highest total likelihood over its lines, and the
scansion of a line is its best path under that meter. Figures~\ref{fig:lattice}--\ref{fig:trellis} show the three
parts on one foot of a line of the poet Fuzûlî, \ot{Kamgir ü}.\footnote{\citet{kilinc2025fuzuli}, p.~315, the
kaside beginning \ot{Ẓulmet-i ḥayretde ẕikrüñdür maña vird-i zebān}, second line of couplet 25.}

\paragraph{Candidate spellings.}
For each plain word $w$, a trained \emph{restorer} returns up to five scholarly
spellings $f$ with probabilities $r(f\mid w)$. The restorer mixes a lexicon of
the spellings written for $w$ in the training editions (weight 0.5, chosen on
development data) with a joint letter $5$-gram over aligned units, each a plain
letter paired with the scholarly letters it stood for \citep{bisani2008joint},
smoothed with Witten--Bell smoothing \citep{witten1991zero} and decoded with a
beam width of 20. On development data the restorer gives \num{RES_right}\% of
\num{RES_words} words the gold spelling. \num{RES_unseen}\% of its errors
are words that do not occur in the training set.

\paragraph{The lattice.}
The restorer gives each word a few candidate spellings along with their
probabilities. However, each spelling can still be read in more than one way,
depending on linking and on medd. For each line we build a lattice that
contains every reading these choices may produce (Figure~\ref{fig:lattice}).
The lattice is a graph whose nodes are the points between syllables and whose
arcs are syllables; a path from the start node to the end node spells out one
reading of the line, syllable by syllable. Given one reading, the weight of
every syllable follows from the rules of \S\ref{sec:aruz}, and the meter model
of the next paragraph scores every path of the lattice against the positions
of a meter.

Each arc carries a label. An open syllable is labeled $\Sw$ and a closed one
$\Lw$, after the short and long positions they normally fill. An overlong
syllable gives two alternative paths between the same nodes, two arcs $\Lw$
and $\Hw$ when it takes medd and one arc $\Ow$ when it does not, as \ot{kām}
does in Figure~\ref{fig:lattice}. A word that ends in a consonant and precedes
a vowel-initial word likewise gives two alternatives, as \ot{gīr} does before
\ot{ü}. In the linked one its final consonant moves to the next word and its
last syllable is re-weighted accordingly (\ot{gī‿rü} $\Lw\Sw$); in the
unlinked one its last syllable is labeled $\Kw$ instead of $\Lw$. $\Ow$ and
$\Kw$ exist because no medd and no linking are flaws. With these labels the
meter model can learn how rare they are, which it could not if they looked
like ordinary closed syllables. Every label is finally paired with the word's
class, Turkish or Arabic/Persian, written as a subscript, $\Lw_{\mathrm{tr}}$
or $\Lw_{\mathrm{ar}}$, so that the model can learn different flaw rates for
the two. A word is Arabic/Persian if its candidate spelling carries a mark Turkish words
do not carry in the Ottoman script. This results in a total of ten labels.

Each candidate spelling of a word contributes one chain of arcs, with the medd
and linking alternatives as parallel branches inside it. For a line of words
$w_1,\dots,w_n$, a path $\pi$ through the lattice selects one spelling $f_i$
for every word, together with one linking and medd choice at each point where
there is one. Its restorer score is the product of the word probabilities,
\begin{equation}\label{eq:theta}
\Theta(\pi)=\prod_{i=1}^{n} r(f_i\mid w_i),
\end{equation}
which acts as a prior over spellings in the joint score of
equation~\ref{eq:joint}; the linking and medd choices carry no prior of their
own. In the lattice, $\log r(f_i\mid w_i)$ is placed on the first arc of the
chain of $f_i$ and every other arc of the chain carries zero, so that the arc
weights along any path sum to $\log\Theta(\pi)$.

\paragraph{The meter HMM.}
The lattice holds every reading of a line. What remains is to decide which
reading is right and under which meter. We do this with a hidden Markov
model whose hidden states are the positions of a meter and whose observed
symbols are the labels of a reading. The first foot of the poem's meter in
Figure~\ref{fig:lattice}, for instance, is the four positions x, \sshort, \slong,
\slong, and the reading \ot{kāmˑ gī‿rü} is explained by these four positions
emitting, in order, the labels $\Lw$, $\Hw$, $\Lw$ and $\Sw$. For each of the
seventeen templates (Appendix~\ref{app:meters}) the model has a left-to-right
chain of \emph{match} states, one per position, of kind long, short or free
(the last position of a line, and the first where the meter allows either, x). A path through a chain is one line of that
template. After each match state there is an \emph{insert} state, which takes
one extra syllable the template has no place for, and over each match state
there is a skip, which passes over one missing position. Neither can be
repeated in a row. Inserts and skips account for lines the meter does not,
such as a misprint in an edition or an error in a copy. The chains are joined
only at \BOS{} and \EOS{}, so a path stays within one template. A long
position emits $\Lw$ most of the time and a short one $\Sw$. The probability
of emitting any other label, which is what a flaw is, is learned from data
and shared by all chains (Figure~\ref{fig:hmm}). In the example, the long
fourth position emits $\Sw$ for \ot{rü}, an imâle. The match, insert and skip
layout is that of a profile HMM \citep{krogh1994hidden}, with emissions
shared across positions of the same kind instead of learned per position.
The probability of a reading under a meter is the probability that the
positions of one of the meter's templates, taken in order, emit the labels of
that reading. Laid against the lattice, the model is a trellis whose rows are
the positions and whose columns are the nodes of the lattice, and a path
through the trellis is one reading placed on the positions
(Figure~\ref{fig:trellis}).

Emissions are tied by the kind of the state, so there are four distributions
over the ten labels, one each for long, short, free and insert states. Every
flaw and choice is one of the rarer emissions and has a rate of its own. Imâle is $\Sw$
emitted by a long position, zihaf is $\Lw$ emitted by a short one, medd is
$\Hw$ emitted by the short position after a long one, no medd is $\Ow$ and no
linking is $\Kw$. Since the labels carry the word's class, each of these rates
is learned separately for Turkish and for Arabic/Persian words.

The model runs over the lattice rather than over a single sequence as in a
standard HMM~\citep{rabiner1989tutorial}. A step pairs one arc of the lattice with one
state of the template, and the state emits the label of the arc. A complete
path pairs one reading $\pi$ of the line with one state sequence $q_{1:T}$
that places its syllables on the positions. With $\sigma_{1:T}$ the labels of
$\pi$, $A$ the transition probabilities and $B$ the emission probabilities,
the score of such a path is
\begin{equation}\label{eq:joint}
\begin{split}
p(\pi,q)=\Theta(\pi)\,A(\BOS,q_1)\prod_{t}B(q_t,\sigma_t)\\
\cdot\Bigl(\prod_{t>1}A(q_{t-1},q_t)\Bigr)A(q_T,\EOS),
\end{split}
\end{equation}
the restorer's prior on the reading times the emissions and transitions along
the path. The probability of a line $x$ under a meter $m$ is the sum of
equation~\ref{eq:joint} over every path of the lattice and every state
sequence within the templates of $m$, computed by the forward algorithm over
the lattice. The meter of a poem is $\argmax_m \log p(m)+\sum_k \log
p(x^{(k)}\mid m)$ over its lines $x^{(k)}$, where $p(x\mid m)$ is normalized by the \BOS{} mass of
$m$'s templates, so the meter prior enters once per poem. Under that meter, the Viterbi
path of each line gives its scansion. The positions come from the states, and
the flaw of each syllable follows from its label and the kind of its state.

Training uses only the printed meter of each poem. A training line is built
into a lattice as above, but from the edition's own spelling, so there is a
single spelling per word and only the linking and medd choices remain open,
and the lattice is scored against the templates of the printed meter alone.
The emission and transition probabilities are then learned by EM~\citep{dempster1977maximum} with
add-$\lambda$ smoothing ($\lambda=0.1$) until the likelihood per line stops
improving. We observed that training everything at once
caused a collapse for the $\Kw$ emission. With the skip transitions available from the start, skipping a
position is cheaper than emitting $\Kw$, so the kept-consonant label never
gathers counts. We therefore train in two stages. The first learns the
emissions with the insert and skip states removed; the second fixes the
emissions of the long, short and free states and learns the transitions, the
meter prior and the insert emissions with the states back in place.

\section{Experimental Setup}\label{sec:baselines}

All trained systems use the training split, are tuned on the development
split of \S\ref{sec:data}, and are evaluated on the test split with the metrics
of \S\ref{sec:tasks}: Task 1 on all \num{NTEST_POEMS} test poems and Task 2
on the 500 annotated test lines. Every system takes as input a poem in plain
letters with no meter given. We compare against four baselines.

\paragraph{Syllable count.} The meter of a poem is the training meter most
frequent for its modal syllable count. This is the simplest baseline for Task 1 and produces no scansion.

\paragraph{Restorer + Kurt \& Kara (2012).} This is the only previous scanner for
Ottoman verse (\S\ref{sec:intro}) that we are aware of. It weighs each syllable of a transcribed
text by rule, picks the meter with the fewest long--short
mismatches over the poem, and labels the mismatches medd, linking, zihaf or imâle by its rules. No code
was released except a web tool which is difficult to use at scale for testing. So we reimplemented their algorithm from the paper. This algorithm runs on scholarly editions, not the modern versions. Thus, we first run the restorer for each word to recover the diacritics and use the result as input to their rule-based algorithm.
We also run the algorithm on the plain letters with no restoration, every vowel counted short, so that the
restorer's share of the result can be read off.

\paragraph{BiLSTM tagger.} A neural sequence labeler~\citep{lample2016neural} trained with the same
supervision as our model, to show what a standard tagger makes of the printed
meter alone. It reads the plain letters of a line character by character,
predicts the poem's meter, and labels each syllable with its position in that
meter; a separate output head restores the spelling. Since no syllable is
labeled in the training data, it is trained to maximize the total
probability of all position sequences that fit the printed meter, which is
the same objective our model's EM optimizes. Linking is applied by rule and
flaws are read off the predicted positions against the syllables' weights. Because a neural model's result varies with its random
initialization, we train five copies with different seeds and report their spread.

\paragraph{Large language models.} We test against Claude Opus 5.5 and
DeepSeek V4.1 Flash, used few-shot. The prompt used explains the task and the
aruz rules of \S\ref{sec:aruz}, lists the fourteen meters with their patterns,
defines the flaws and linking, fixes the output format and gives six scanned
development lines as examples. The full prompt is in the released code. Each
model answers once per poem in JSON, with reasoning enabled. An answer that
does not parse, or whose positions fit none of the
meter's patterns, is counted wrong throughout (1 of \num{FLASH_asked} answers
for Opus 5.5, 66 for Flash). Opus 5.5 was one of the three annotators of
\S\ref{sec:data} to produce the gold. DeepSeek V4.1 Flash was chosen as an
independent lineage.

\paragraph{Ablations.} To see which parts of the model account for its
result, we retrain and evaluate it with one part removed at a time. Without
the HMM, the fewest-flaw rule of Kurt \& Kara chooses among the restorer's
five candidates. Without EM, the model keeps its starting values. Without the
restorer's probabilities, its candidates enter the lattice with a flat prior,
and without the restorer altogether, the candidates come from the training
lexicon. Without the labels $\Ow$ and $\Kw$, no medd and no linking are
emitted as plain closed syllables. Finally, we give the model the
edition's scholarly transcription in place of plain letters, akin to an oracle restorer.

\section{Results}\label{sec:results}

Table~\ref{tab:main-test} gives the main results on the test split. Every system that restores the spelling
identifies the meter of nearly every poem; ours and Opus 5.5 are above 99.8\% per poem, restorer + Kurt \& Kara at
\num{KurtKararestorer_t1poem}\%, and even the plain-letter run of Kurt \& Kara reaches \num{KKPLAIN_poem}\%, because a wrong template mismatches a whole
poem more than the right one does. The systems separate per line, where ours reaches \num{OURS_t1line}\% against
\num{KurtKararestorer_t1line}\% for restorer + Kurt \& Kara and \num{CLF_line}\% for the tagger. This shows that
meter identification itself is not a hard task. The real task of this work is scansion.
\begin{table}[t]\centering\small\setlength{\tabcolsep}{3pt}
\resizebox{\columnwidth}{!}{%
\begin{tabular}{@{}lrrrrr@{}}\toprule
 & \multicolumn{2}{c}{Task 1: meter} & \multicolumn{3}{c}{Task 2: scansion}\\
\cmidrule(lr){2-3}\cmidrule(lr){4-6}
system & poem & line & full & flaw F1 & syll.\ F1\\\midrule
syllable count & 56.6 & 57.2 & -- & -- & --\\
BiLSTM tagger (5 seeds) & 97.8 & 92.0 & 55.0 & 80.4 & 96.0\\
Kurt \& Kara, plain letters & 91.0 & 71.4 & 2.4 & 42.0 & 85.8\\
restorer + Kurt \& Kara & 98.9 & 84.1 & 26.8 & 54.9 & 90.5\\
\textbf{ours}: HMM + restorer & \textbf{99.9} & \textbf{98.2} & \textbf{77.6} & \textbf{89.6} & \textbf{98.0}\\
Claude Opus 5.5 & 99.9 & -- & 63.6 & 83.9 & 95.1\\
DeepSeek V4.1 Flash & 93.1 & -- & 40.0 & 70.3 & 84.8\\
\bottomrule
\end{tabular}}
\caption{Test results on plain Turkish letters with no meter given. Task 1: meter accuracy per poem and per line, on all test poems. Task 2: full-line accuracy (full), pooled flaw F1 and syllable F1, on the 500 annotated test lines. The tagger is the mean of five seeds. A language model answers once per poem and has no per-line meter.}\label{tab:main-test}
\end{table}

Our model reads \num{OURS_exact}\% of the annotated test lines exactly. The nearest system is Opus 5.5 at
\num{OPUS_exact}\%, a difference of \num{SIG_opus_diff} points ($\num{SIG_opus_ptxt}$, 95\% CI
\num{SIG_opus_lo} to \num{SIG_opus_hi}).\footnote{Exact McNemar test~\citep{mcnemar1947note} over the 500 test lines,
on the lines where the two systems disagree, with a paired bootstrap confidence interval over the same lines. The
tagger is compared seed by seed.} The
tagger reads \num{TAG_exact}\% on average and is behind ours on every seed, and Flash reads \num{FLASH_exact}\%. Restorer + Kurt \& Kara
reads \num{KurtKararestorer_exact}\%, and the same algorithm on the
plain letters only \num{KKPLAIN_exact}\%, every long vowel having become an open syllable in a long position. Flaw F1
ranks the systems in the same order. These results show that our model, built for the task, is better
than out-of-the-box large language models and the previous rule-based scanner. Being a generative model, it also gives
a probability for every decision it makes, the posterior of a reading under the meter, whereas a language model's
answer comes with no reliable measure of its own confidence~\citep{xiong2024can}.

\paragraph{Ablations.}
Table~\ref{tab:ablation-test} contains the ablation results. No single part of the model accounts for its results. Without
the HMM, the fewest-flaw rule over the same lattice of five candidates that the HMM receives reads \num{ABL_nohmm} points fewer lines; without the
restorer's probabilities, \num{ABL_flat} fewer; without the labels $\Ow$ and $\Kw$, \num{ABL_nosymbols} fewer.
Each of these leaves the model with less than half of its fully read lines, so the restorer's probabilities, the scoring under
the meter and the labels for the two flaws that the letters hide are each necessary, and the gain over the
baselines comes from deciding them together. A scanner
for aruz has to weigh the spelling behind the plain letters, the flaws the poet
took and the meter of the poem against one another, since none of them is
certain on its own. The structure matters more than the learned values, since keeping
the starting values in place of EM costs \num{ABL_noem} points, and the restorer's generalization to unseen words
matters, since the training lexicon in its place costs \num{ABL_lexicon} points. Finally, with the edition's scholarly transcription in place of the restorer's
candidates, the model reads \num{SCH_HMM}\% of lines. Of its \num{ERR_total} points of error on plain letters, \num{ERR_spelling} are
therefore in the spelling and \num{ERR_rest} in linking, medd and the flaw labels, which the model gets wrong even when the
spelling is given. Kurt \& Kara reads \num{KKSCH_exact}\% of lines from the same scholarly transcription, so most of the
distance between the two systems is in the scansion rules, not in the spelling. A user who trusts a scholarly transcription can therefore give it to the tool
as it is, with its diacritics kept, and read \num{SCH_HMM}\% of lines rather
than \num{OURS_exact}\%; the web tool offers both inputs.

\begin{table}[t]\centering\small\setlength{\tabcolsep}{3pt}
\resizebox{\columnwidth}{!}{%
\begin{tabular}{@{}lrrrrr@{}}\toprule
 & \multicolumn{2}{c}{Task 1: meter} & \multicolumn{3}{c}{Task 2: scansion}\\
\cmidrule(lr){2-3}\cmidrule(lr){4-6}
ablation & poem & line & full & flaw F1 & syll.\ F1\\\midrule
full model & 99.9 & 98.2 & 77.6 & 89.6 & 98.0\\
no HMM (aligner) & 99.9 & 96.3 & 33.4 & 56.6 & 95.7\\
no EM & 96.8 & 85.0 & 59.4 & 77.1 & 94.7\\
no O/K symbols & 99.9 & 97.4 & 35.2 & 64.7 & 91.0\\
flat candidate probs. & 99.7 & 95.1 & 28.4 & 53.6 & 94.5\\
lexicon, no restorer & 99.6 & 94.2 & 37.6 & 73.2 & 92.4\\\midrule
ours, scholarly text & 99.9 & 98.9 & 90.6 & 94.8 & 98.5\\
Kurt \& Kara, scholarly text & 99.1 & 84.9 & 34.4 & 59.8 & 90.8\\
\bottomrule
\end{tabular}}
\caption{Ablations on test, with one part of the model removed per row (\S\ref{sec:baselines}). Columns as in Table~\ref{tab:main-test}. The last two rows give our model and Kurt \& Kara the edition's scholarly transcription instead of the plain letters. The insert and skip states are ablated in Figure~\ref{fig:robustness}.}\label{tab:ablation-test}
\end{table}

\paragraph{Error analysis.}
A wrong line usually differs from the reference in more than one place. In our error analysis, we
count each line once, under the first of these differences in a fixed order:
the meter; the spelling of a word, when the restored form differs in its
consonants; its vowel length, when it differs only in a long or short vowel;
a medd; a linking; and the flaw label, when form, linking and positions all
match and only the label does not. The last can happen only to the large language
models, since for the other systems the label follows from the form and the
position. Our model's \err{ours_total} wrong lines are
mostly vowel lengths (\err{ours_vowellength}) and linkings (\err{ours_linking}); spellings (\err{ours_spelling})
and medd (\err{ours_medd}) account for the rest. The wrong vowel lengths are the restorer's errors the meter could
not overrule, because both readings fit a position. The wrong linkings are lines where the reference keeps a final consonant that
the model links. They follow from what the model has learned: $\Kw$, the label of a word-final
consonant that is not linked to the following vowel, is emitted by a long position with probability
\num{EM_K_long_tr} for Turkish and \num{EM_K_long_ar} for Arabic or Persian
words, so the model keeps a consonant only where the meter forces it.

The other learned rates read the same way as a description of the poets' practice (Appendix~\ref{app:learned}).
Imâle is common and falls on Turkish words, a Turkish syllable in a long position being open \num{EM_imale_tr}\%
of the time against \num{EM_imale_ar}\% for an Arabic or Persian one, and zihaf is rare, \num{EM_zihaf_tr}\% and
\num{EM_zihaf_ar}\% of syllables in short positions. These rates agree with the qualitative account of the standard aruz
philology~\citep{duzenli2018aruz}, although they were learned from the data alone.

\begin{figure}[t]\centering
\includegraphics[width=\columnwidth]{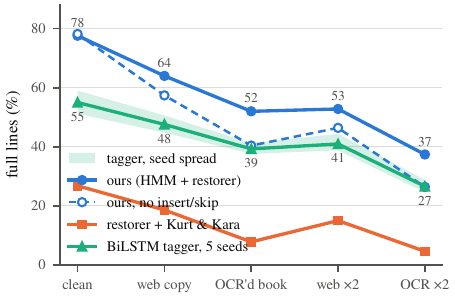}
\caption{Full-line accuracy on the test set, for the
clean test poems, their web copy and OCR'd book at the measured deviation
rates (\S\ref{sec:data}), and the same at twice every rate.}
\label{fig:robustness}
\end{figure}

\paragraph{Noisy copies.}
The robustness benchmark of \S\ref{sec:data} rewrites the test poems as the
copies a user is likely to paste, once as a web page and once as an OCR'd book, with the
kinds and rates of deviation measured on real copies, and once more at twice
every rate. Figure~\ref{fig:robustness} gives full-line accuracy on each. A
web copy reduces the performance of our model by \num{COSTWEB}\% of its fully correct lines and an
OCR'd book \num{COSTOCR}\%, while per-line meter accuracy stays at
\num{NOISEMETER}\%. The meter is decided from the whole poem, so the unaffected lines carry it, and the correct meter in turn helps the scansion of the affected lines. The tagger drops at a similar rate, while restorer + Kurt \& Kara loses most of its correct lines (\num{ROB_web_KK_exact}\% on the web copy, \num{ROB_ocr_KK_exact}\% on the OCR'd book). Our model without the insert/skip states (Figure~\ref{fig:robustness}) falls below the performance of comparable models for some perturbations, justifying the need for the insert/skip states for the robustness of the models.

On every noisy input ours has the best full-line accuracy, \num{DMIN} to \num{DMAX} points
ahead of the tagger on all five seeds on every input. However, the perturbations affect all models very clearly and the performance starts to fall to levels that are not suitable for real-world use even before the rates are doubled, motivating further work in robustness of scansion models.

In the errors of our model, modernized
spelling as part of the perturbations (\ot{meni} → \ot{beni}) alone costs \num{FORMSLOSS} points although it changes no weight, because a modernized word is one the restorer has not seen. The importance of the restorer model is apparent in this case too.

The model's score also says how well a line fits its meter, which can be used to flag the lines a copy has
corrupted. A detector built on the per-syllable log-likelihood, the insert states and a leave-one-out test over the
words, thresholded on clean development lines at 5\% false alarms, catches \num{DET_web_rec}\% of the lines whose
weights a web copy changed at \num{DET_web_prec}\% precision, and \num{DET_ocr_rec}\% at \num{DET_ocr_prec}\% on
the OCR'd book (Table~\ref{tab:detection-test},
Appendix~\ref{app:robustness}).

\section{Related Work}\label{sec:related}

Automatic scansion in NLP has been developed mainly for European verse, as
sequence labeling of stress, with rule-based scanners
\citep{navarro2018metrical,delarosa2020rantanplan}, HMMs, CRFs and neural
taggers \citep{agirrezabal2016machine,agirrezabal2017comparison}, and lately
fine-tuned encoder--decoders \citep{ruizfabo2026scansion}. For the quantitative meters of
Arabic, Persian and Sanskrit, including aruz, work has concentrated on identifying the meter,
by matching the syllable pattern against the inventory
\citep{abuata2018rule,terdalkar2022chandojnanam} or by classifying it from the
characters with a neural model
\citep{yousef2019learning,alshaibani2020meter,alyafeai2023ashaar,mutawa2025determining} with relatively little attention to scansion alignment and noisy copies. For Ottoman Turkish
the only system we found is the rule-based scanner of \citet{kurt2012arud}, which reads
syllable weights off a hand-made scholarly transcription and counts
mismatches. The handbooks
\citep{ipekten2010aruz,sarac2007bicim}, \citet{cetin1991aruz} and
\citet{duzenli2018aruz} describe the flaws qualitatively. 

Restoring what an orthography omits has been studied as diacritic restoration
\citep{mihalcea2002letter,belinkov2015arabic,adali2014vowel}, with the restored
form as the end product. Our restorer is a joint-sequence model of the kind
built for grapheme-to-phoneme conversion \citep{bisani2008joint}. Here the
restored form is one of several readings a line admits, and which is right
depends on a meter that is itself unknown, so restoration and scansion are one joint
search, a lattice of readings decoded by an HMM \citep{rabiner1989tutorial}
with the match, insert and skip structure of a profile HMM
\citep{krogh1994hidden,durbin1998biological}.

\section{Conclusion}

We addressed the scansion of Ottoman Divan poetry mainly from the plain modern
spelling, with capabilities on scholarly editions as well. We built a corpus of
52,000 lines with scansion annotations for 500 test lines, a robustness
benchmark from the deviation rates measured on web and OCR copies, and a model
that decides spelling, linking and position together, trained from the printed
meter of each poem alone. On held-out poets it reads \num{OURS_exact}\% of
lines fully correctly, ahead of the comparable models.
The ablations and error analysis show that joint decoding is critical for this task. Spelling, linking and position are each uncertain on their own
and also reliant on the others. Systems that fix any one of them before
the rest tend to perform worse. The learned flaw rates agree with
the philological account without annotation. We open-source the code and release the scansion tool on the web.

\section*{Limitations}

The scansion annotation was made by an ensemble of large language models under a written guide,
with agreement measured and 200 lines hand-checked. A wider test set that is hand annotated by multiple experts would be helpful in further error analysis with the hopes of producing better and more robust models. The inventory covers fourteen meters; \num{NTEST_OUT} of the \num{NTEST_ALL}
test poems fall outside it and are not scored, and the model does not report
that a poem fits no meter. The robustness benchmark is generated from measured
deviation rates, not from the real copies, whose transcriptions may be
protected. The benchmark is also released as a pipeline rather than as a full dataset, as the editions of the poems are protected.

\bibliography{paper,robustness}

\appendix
\setcounter{topnumber}{4}\setcounter{dbltopnumber}{4}\setcounter{totalnumber}{6}
\renewcommand{\topfraction}{.95}\renewcommand{\dbltopfraction}{.95}
\renewcommand{\textfraction}{.02}\renewcommand{\floatpagefraction}{.8}\renewcommand{\dblfloatpagefraction}{.8}

\section{The Meter Inventory}\label{app:meters}

Table~\ref{tab:meters} lists the \num{N_METERS} meters of the inventory with their \num{N_TEMPLATES} templates (a
meter whose first or last foot has two accepted forms contributes one template for each), and the share of training
poems and of training lines in each; \num{N_TRAIN_INVENTORY} of the 1,644 training poems are in the inventory.
\begin{table*}[t]\centering\footnotesize\setlength{\tabcolsep}{4pt}
\begin{tabular}{@{}llrrr@{}}\toprule
meter, as the editions print it & template(s) & $n$ & train poems \% & train lines \%\\\midrule
\ot{fāʿilātün fāʿilātün fāʿilātün fāʿilün} & \pat{x.---.---.---.-} & 1 & 31.4 & 30.1\\
\ot{mefāʿīlün mefāʿīlün mefāʿīlün mefāʿīlün} & \pat{.---.---.---.---} & 1 & 19.8 & 18.7\\
\ot{mefʿūlü fāʿilātü mefāʿīlü fāʿilün} & \pat{--.-.-..--.-.-} & 1 & 13.0 & 11.5\\
\ot{mefʿūlü mefāʿīlü mefāʿīlü feʿūlün} & \pat{--..--..--..--} & 1 & 11.1 & 8.7\\
\ot{feʿilātün feʿilātün feʿilātün feʿilün} & \pat{x.--..--..----}, \pat{x.--..--..--..-} & 2 & 9.7 & 14.0\\
\ot{mefāʿilün feʿilātün mefāʿilün feʿilün} & \pat{.-.-..--.-.---}, \pat{.-.-..--.-.-..-} & 2 & 6.1 & 5.0\\
\ot{mefāʿīlün mefāʿīlün feʿūlün} & \pat{.---.---.--} & 1 & 3.0 & 3.3\\
\ot{fāʿilātün fāʿilātün fāʿilün} & \pat{x.---.---.-} & 1 & 2.1 & 2.7\\
\ot{feʿilātün mefāʿilün feʿilün} & \pat{x.--.-.---}, \pat{x.--.-.-..-} & 2 & 1.9 & 3.9\\
\ot{mefʿūlü fāʿilātün mefʿūlü fāʿilātün} & \pat{--.-.----.-.--} & 1 & 0.8 & 0.6\\
\ot{müstefʿilün müstefʿilün müstefʿilün müstefʿilün} & \pat{--.---.---.---.-} & 1 & 0.6 & 0.7\\
\ot{mefʿūlü mefāʿilün feʿūlün} & \pat{--..-.-.--} & 1 & 0.3 & 0.4\\
\ot{feʿūlün feʿūlün feʿūlün feʿūl} & \pat{.--.--.--.-} & 1 & 0.1 & 0.1\\
\ot{müfteʿilün müfteʿilün fāʿilün} & \pat{-..--..--.-} & 1 & 0.1 & 0.2\\
\bottomrule
\end{tabular}
\caption{The meter inventory. A meter whose first or last foot has two accepted forms has two templates; x is a position that may be long or short, and the last position of every template, printed long, is free. Shares are of the training poems and lines in the inventory.}\label{tab:meters}
\end{table*}

\section{Learned Values}\label{app:learned}

Table~\ref{tab:emis} gives the emission distributions of the trained model, one row per state kind, over the ten
labels.
\begin{table*}[t]\centering\footnotesize\setlength{\tabcolsep}{4pt}
\begin{tabular}{@{}lrrrrrrrrrr@{}}\toprule
position kind & L:tr & S:tr & H:tr & O:tr & K:tr & L:ar & S:ar & H:ar & O:ar & K:ar\\\midrule
long $\beta_\Lw$ & .362 & .141 & $7{\cdot}10^{-7}$ & $7{\cdot}10^{-7}$ & $3{\cdot}10^{-4}$ & .429 & .066 & $7{\cdot}10^{-7}$ & $5{\cdot}10^{-6}$ & .002\\
short $\beta_\Sw$ & $2{\cdot}10^{-4}$ & .571 & .006 & $10^{-6}$ & $10^{-6}$ & .005 & .383 & .035 & $10^{-6}$ & $10^{-6}$\\
free $\beta_\Xw$ & .434 & .157 & 0 & $4{\cdot}10^{-6}$ & $3{\cdot}10^{-4}$ & .352 & .056 & 0 & $4{\cdot}10^{-6}$ & .001\\
insert $\beta_\Iw$ & .014 & .071 & .014 & .014 & .014 & .026 & .579 & .240 & .014 & .014\\
\bottomrule
\end{tabular}
\caption{Learned emission probabilities, one row per state kind, over the ten labels (L closed, S open, H the half beat of a medd, O an overlong syllable in one position, K an unlinked final consonant; tr Turkish, ar Arabic/Persian).}\label{tab:emis}
\end{table*}

\section{The Robustness Benchmark}\label{app:robustness}

\paragraph{The real copies.} A page or a scanned book counts as a copy of a corpus poem when two of its lines match
corpus lines at a character similarity of 0.75 or one at 0.8, after line numbers, trailing citations and footnote
digits are stripped. The web copies are \num{ROBX_web_sources} documents with at least fifty words
(\num{ROBX_web_survey_misras} lines, \num{ROBX_web_survey_words} edition words), mostly a user anthology site, with
school sites, listicles, blogs and a music repertoire; the OCR copies are \num{ROBX_ocr_sources} books on
archive.org (\num{ROBX_ocr_survey_misras} lines, \num{ROBX_ocr_survey_words} words): two scanned Latin-script
editions of Fuzûlî, a commentary, a thesis and two popular books. One poet dominates the copies, the development
poets have none, and no copy typed without the Turkish letters was found. Within a matched line the copy's words are
aligned to the edition's by edit distance, allowing splits and joins. Both sides first pass through the systems'
input normalizer (length and consonant marks, ayn, hemze and apostrophes dropped, hyphens joined), so a copy that
drops every circumflex differs from the edition in nothing the benchmark counts; Table~\ref{tab:rob-catalogue} is
what survives that fold.
\begin{table*}[t]\centering\footnotesize\setlength{\tabcolsep}{4pt}
\begin{tabular}{@{}lrrrrl@{}}\toprule
deviation (after the alphabet fold) & anthology & OCR edition & OCR popular book & changes weights & example\\\midrule
vowel changed in the ending & 4.08 & 9.65 & 9.33 & 0 & \ot{istersen $\to$ istersin}\\
Turkish letter lost or swapped & 1.73 & 5.63 & 1.72 & 0 & \ot{zulmı $\to$ zulmi}\\
letter edit, non-word & 1.55 & 3.75 & 5.16 & 28 & \ot{çekmekden $\to$ çekmekten}\\
letter edit, another word & 2.78 & 2.86 & 3.38 & 17 & \ot{dideyi $\to$ didei}\\
vowel changed & 1.33 & 2.57 & 2.84 & 1 & \ot{ıkdı $\to$ akdi}\\
vowel changed in the stem & 2.16 & 1.34 & 2.37 & 0 & \ot{etseler $\to$ itseler}\\
word substituted & 1.01 & 1.24 & 0.96 & 41 & \ot{kılam $\to$ edeyüm}\\
word split & 0.65 & 0.94 & 0.57 & 52 & \ot{bilseydüm $\to$ bilse idüm}\\
men/maña written ben/bana & 1.33 & 1.04 & 2.16 & 0 & \ot{menüm $\to$ benüm}\\
words joined & 0.36 & 0.86 & 0.66 & 59 & \ot{u perişan $\to$ perişan}\\
h written k & 0.54 & 0.81 & 0.96 & 0 & \ot{yohsa $\to$ yoksa}\\
vowel inserted & 0.43 & 0.40 & 0.71 & 100 & \ot{aşnamuz $\to$ aşinamız}\\
long vowel typed double & 0.07 & 0.34 & 0.55 & 100 & \ot{efgan $\to$ efgaan}\\
word dropped & 0.11 & 0.25 & 0.22 & 100 & \ot{dem $\to$}\\
word added & 0.04 & 0.23 & 0.11 & 93 & \ot{$\to$ mülki}\\
vowel dropped & 0.04 & 0.24 & 0.08 & 100 & \ot{ede $\to$ de}\\
izafet dropped & 0.07 & 0.13 & 0.05 & 100 & \ot{garibi $\to$ garib}\\
final devoicing & 0.07 & 0.09 & 0.00 & 0 & \ot{azad $\to$ azat}\\
double consonant added & 0.07 & 0.03 & 0.03 & 65 & \ot{rab $\to$ rabb}\\
double consonant dropped & 0.00 & 0.03 & 0.00 & 100 & \ot{muarrif $\to$ muarif}\\
\bottomrule
\end{tabular}
\caption{Deviations found in real copies, per 100 edition words, by source; ``changes weights'' is the share of that deviation's words whose syllable weights change.}\label{tab:rob-catalogue}
\end{table*}
\begin{table*}[t]\centering\footnotesize\setlength{\tabcolsep}{4pt}
\begin{tabular}{@{}lrrrrrr@{}}\toprule
 & lines & dev./w. & letters & syll. & weights & CER\\\midrule
web copies, measured & 512 & 20.6 & 77.3 & 6.8 & 18.6 & 4.8\\
\quad generated, $\times$1 & 14,516 & 19.5 & 72.1 & 5.1 & 18.1 & 5.2\\
\quad generated, $\times$2 & 13,018 & 25.6 & 81.7 & 9.2 & 30.6 & 7.7\\
OCR'd books, measured & 13,532 & 28.9 & 88.6 & 10.3 & 23.4 & 6.3\\
\quad generated, $\times$1 & 15,992 & 27.6 & 85.8 & 13.6 & 25.0 & 7.3\\
\quad generated, $\times$2 & 15,992 & 42.2 & 95.8 & 24.0 & 42.9 & 12.4\\
\bottomrule
\end{tabular}
\caption{The real copies' severity against the generated files', on the survey's measures in the standard Turkish alphabet: word deviations per edition word, lines whose letters, syllable count or syllable weights differ from the edition's, character error rate (percent). The generator's rates are scaled by one factor for the edits that keep the weights and one for those that change them so that the full profile reaches the measured word deviations per word and the measured share of lines whose weights differ; the other measures are not fitted.}\label{tab:rob-severity}
\end{table*}
\begin{table*}[t]\centering\footnotesize
\begin{tabular}{lrrrrrrrr}\toprule
input & faulty lines & AUROC & precision & recall & alarms, scansion kept & top-1 word & kind & word offered\\\midrule
web & 618/3438 & 0.714 & 51.3 & 42.4 & 10.1 & 57.0 & 83.9 & 3.4\\
web x2 & 888/3038 & 0.724 & 62.3 & 47.0 & 13.5 & 53.2 & 83.9 & 4.2\\
web forms & 0/3742 & -- & -- & -- & 10.0 & -- & -- & --\\
web damage & 632/3742 & 0.711 & 57.9 & 40.5 & -- & 60.7 & 90.8 & 4.4\\
web clutter & 0/3742 & -- & -- & -- & -- & -- & -- & --\\
web structure & 0/3428 & -- & -- & -- & -- & -- & -- & --\\
ocr & 955/3742 & 0.774 & 66.5 & 54.3 & 10.3 & 55.4 & 89.7 & 22.3\\
ocr x2 & 1552/3742 & 0.791 & 75.8 & 61.9 & 14.6 & 49.8 & 85.6 & 21.0\\
ocr forms & 0/3742 & -- & -- & -- & 9.5 & -- & -- & --\\
ocr damage & 845/3742 & 0.742 & 68.4 & 44.3 & -- & 64.6 & 91.9 & 35.2\\
ocr clutter & 120/3742 & 1.000 & 34.5 & 100.0 & -- & 27.5 & 100.0 & --\\
\bottomrule\end{tabular}
\caption{The warning detector on each noisy input. A line is faulty if its syllable weights differ from the edition's. AUROC, and precision and recall at the threshold set on clean development lines; alarms on lines whose scansion did not change; and, among warned faulty lines, how often the named word is faulty, the stated kind of damage is right, and the edition's word is among those offered.}
\label{tab:detection-test}\end{table*}

\paragraph{Producing the copies.} Every letter-level edit seen in an aligned word pair is kept with one letter of context
on its left and two on its right, and its rate is the number of times it was made over the number of times its
context occurs in the edition words of the same copies; an edit enters the table when seen at least twice (web
\num{ROBX_web_wordedits} edits, OCR \num{ROBX_ocr_wordedits}). It counts as damage when it changed the word's
weights in at least half of its occurrences, else as forms. One table thus covers modernization, inserted and
doubled vowels, OCR confusions and typing errors, each at its own rate. Substitutions draw an observed triple
(letter similarity, syllable-count difference, weights kept or changed) and a training-vocabulary word that matches
it; dropped izafets, dropped, added, split and joined words, line numbers, footnote digits and citations (clutter) occur at
their measured rates and in their observed styles; on the web profile a couplet is left out (structure) with probability
\num{ROBX_web_beyit_left_out}\%. Each poem draws one intensity from the observed per-document rates
(\num{ROBX_web_intensity_min} to \num{ROBX_web_intensity_max} times the web profile), since real sources differ in how
much they deviate~\citep{naplava2021understanding}, and one copy deviates at one rate throughout. Two factors per profile, one on the
forms rates and one on the damage rates (web \num{ROBX_web_factor_forms} and \num{ROBX_web_factor_damage}, OCR
\num{ROBX_ocr_factor_forms} and \num{ROBX_ocr_factor_damage}), bring the generated copy close to the measured word
deviations per word and share of lines whose weights differ; the web forms factor is large because the web survey is
small and only edits seen twice enter the table. Table~\ref{tab:rob-severity} compares the generated files with the
real copies on these two measures and on three that are not fitted. Every system receives the same word list from
a copy. The text goes through the normalizer, is split on spaces and stripped of punctuation, and tokens without a
vowel are dropped. A line number is therefore invisible to all systems, a footnote digit glued to a word is seen by
the restorer's lookup and by the tagger, and a citation adds words.

\paragraph{The warning detector.} Table~\ref{tab:detection-test} gives the detector of \S\ref{sec:results} on
each noisy input.

\end{document}